\documentclass{article}

\PassOptionsToPackage{numbers,sort&compress}{natbib}

\usepackage{graphicx}

\usepackage[sglblindworkshop, final]{neurips_2026}

\usepackage[utf8]{inputenc} 
\usepackage[T1]{fontenc}    
\usepackage{hyperref}       
\usepackage{url}            
\usepackage{booktabs}       
\usepackage{amsfonts}       
\usepackage{nicefrac}       
\usepackage{microtype}      
\usepackage{xcolor}         
\usepackage{booktabs}
\usepackage{tabularx}

\title{Does Pretraining Teach\\Chemical Language Models Chemistry?}
\workshoptitle{AI4Science}

\author{%
  Leif Sieben\thanks{Both authors contributed equally, names listed in alphabetical order.} \\
Independent Researcher \\
Oxford, UK \\
\texttt{leif.sieben@gmail.com} \\
   \And
   Yoel Zimmermann$^*$ \\
   Harvard University \\
   Cambridge, MA, USA \\
   \texttt{yzimmermann@g.harvard.edu} \\
}

\begin{document}

\maketitle

\begin{abstract}
Chemical language models (CLMs) are pretrained on large corpora of molecules with the aim of learning representations that transfer to diverse downstream tasks. Here, we systematically study the effects of pretraining using a series of such CLMs across 64 molecular property prediction datasets. We find that the downstream gains of pretraining in CLMs do not scale and performance saturates below gradient-boosting baselines. Across the regimes evaluated, only supervised pretraining on traditional physico-chemical descriptors consistently improves downstream performance for the CLMs tested whereas some forms of pretraining result in models worse than a randomly initialized encoder. While models can resolve molecular structure well, they do so no better than a randomly initialized model, suggesting that, at best, CLMs learn descriptor-like properties in addition to low-level token statistics during pretraining, but surprisingly little chemistry beyond what classical molecular features already capture. 
\end{abstract}

\section{Introduction}

The pretraining of models on a diverse corpus before adapting them to a downstream task of interest by fine-tuning has been a key paradigm in the development of modern machine learning. The pretraining paradigm naturally motivates the training of very large foundation models that often outperform specialist models, which only see data from the specific task they are evaluated on~\cite{Bommasani2021On, devlin2019bert, kornblith2019better}. The benefits of pretraining are generally understood to arise from the acquisition of reusable structure from the underlying domain. In natural language, pretraining has been shown to produce representations that encode linguistic structure, including parts of speech, syntax, and semantic information~\cite{tenney2019bert, liu2019linguistic, rogers2020primer}. Similarly, vision models learn to encode semantic and object-level structure despite receiving no corresponding labels~\cite{caron2021emerging}.

Given its success in other domains, the pretraining paradigm has also been applied to chemistry. Treating molecules as objects for machine learning requires them to be converted into suitable mathematical representations, which can be done via hand-crafted rules, notably descriptors, which are computed values describing specific molecular properties \cite{todeschini2000handbook,todeschini2009molecular}, through molecular fingerprints \cite{rogers2010ecfp, morgan1965generation}, or, increasingly, by using the embeddings of machine learning models \cite{duvenaud2015convolutional, david2020molecular}. One common class of such models is trained analogously to language models, where instead of taking text as input these models are trained on molecules formatted as strings using SMILES~\cite{weininger1988smiles}. Models can be trained fully unsupervised (i.e., using only unlabeled data) by making use of masked language modeling (MLM) \cite{devlin2019bert}, where the model is tasked with filling in missing tokens in a string. We refer to this type of model as a ``chemical language model (CLM)'', which is equivalent to another commonly used term, ``molecular language models'', with the advantage of a more distinct acronym. A CLM is trained on molecular data alone (i.e., structures and optionally molecular properties) as opposed to LLMs that are fine-tuned or entirely trained on natural language corpora on the subject of chemistry (such as scientific literature, patents or experimental records)~\cite{zhang2024scientificlargelanguagemodels} or general-purpose LLMs used for chemical tasks~\cite{jablonka2024leveraging}.

In this work, we focus on the use of CLMs for molecular property prediction, where a model is tasked with predicting a property such as solubility, biological activity, or quantum-mechanical energy. We ask: does pretraining produce representations that encode chemically relevant information transferable to such downstream tasks? We illuminate this question from three perspectives: whether pretraining improves downstream prediction and continues to improve with the amount of pretraining data; whether these gains depend on chemically meaningful relationships in the pretraining data; and whether the resulting representations contain information beyond established molecular features and resolve chemically meaningful changes in molecular structure.

\paragraph{Related Work}

Prior work has produced mixed evidence that pretraining improves molecular property prediction. For GNNs, Sun, Dai, and Yu~\cite{sun2022gnnpretrain} found substantially larger gains from supervised than unsupervised pretraining, with weak scaling in unlabeled data and little benefit when relevant descriptors were available. Similar results have emerged for CLMs: MoLFormer showed no clear downstream scaling with pretraining dataset size~\cite{ross2022molformer}, while ChemBERTa-2 found supervised multi-task regression (MTR) generally superior to MLM and that improved pretraining loss did not reliably translate to downstream gains~\cite{ahmad2022chemberta2}. Sultan \textit{et al.} found combined MLM+MTR optimal but still below downstream-trained tree models~\cite{sultan2025transformersmolecularpropertyprediction}, while Sagawa and Kojima found that downstream performance resists improvement even as pretraining loss scales with model and dataset size as well as compute~\cite{sagawa2026largescalechemicallanguagemodels}.

Another line of work has asked what molecular information CLMs acquire during pretraining. Probing and interpretability studies have found evidence that pretrained representations encode aspects of molecular syntax, substructures, and property-relevant features~\cite{ross2022molformer, kenneth2026does}.
Karnysheva \textit{et al.}~\cite{karnysheva2026probingchemicallanguagemodels} found that molecular substructures relevant for a downstream task undergo larger representational changes during fine-tuning than less relevant ones, but leave open whether this results in improved downstream property prediction. Interestingly, even randomly initialized models already encoded the basic motif of ring structures.

Our work connects these two lines of research by asking not only whether pretraining improves molecular property prediction, but what information is responsible for any resulting improvement. Across systematic supervised and unsupervised pretraining experiments, we test how downstream performance changes with pretraining dataset size, molecular and task similarity, and across control experiments where we pretrain on increasingly perturbed input data to isolate what information is required for successful pretraining. We further investigate whether the resulting representations contain information beyond established molecular fingerprints and descriptors and whether they resolve chemically meaningful changes in molecular structure. Together, our experiments provide tests of whether the benefits of molecular pretraining can be attributed to learning transferable chemical information.

\section{Results}

A foundation model in chemistry should generate informative embeddings of molecules that only require the fine-tuning of small prediction heads, such as a multi-layer perceptron (MLP) for property prediction. In chemistry, we are privileged to have access to a fast, robust, and powerful alternative to such embeddings by using fingerprints, optionally supplemented with physico-chemical descriptors. We thus benchmark our CLM series of models, referred to as \textbf{C}hemical \textbf{L}anguage \textbf{I}nference with \textbf{M}odern \textbf{B}ERT (CLIMB), against these baseline encodings as input to XGBoost \cite{chen2016xgboost}, a gradient boosting algorithm that is generally considered a very strong baseline. We evaluated these models on a total of 64 datasets using three independent model seeds evaluated on five-fold cross-validation on each dataset. We report performance split among four main types of tasks: classification and regression, as well as activity cliffs (where a small structural change has a large property-level effect) and virtual screenings (which are large, heavily skewed classification problems, where few actives must be found among many inactive molecules). A list of the datasets used can be found in SI section \ref{sec:downstream_datasets}. We trained most CLIMB models on a corpus of 8M molecules which allowed us to train more models, including the three seed replicates. We also include the results for two scaled-up models pretrained on 100M unique molecules which, due to compute limitations, only have one seed each. 

\begin{figure}[t]
    \centering
    \includegraphics[width=\linewidth]{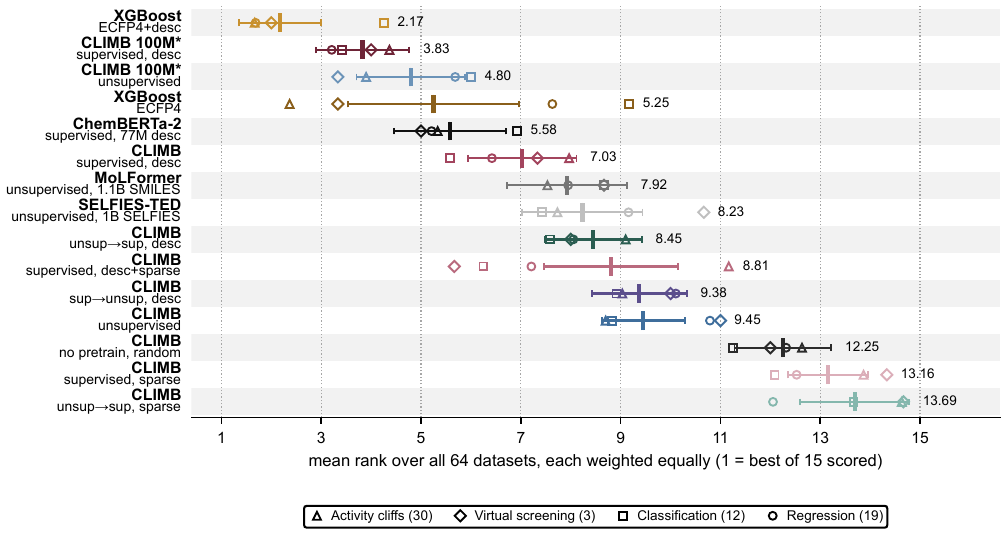}
    \caption{\textbf{Overall benchmarking performance.}
    Mean rank across 64 datasets grouped into four task families. ($*$) All CLIMB models use three seeds, except the 100M models at one seed.}    
    \label{fig_A}
\end{figure}

The CLIMB models cover a surprisingly wide range of performance and some do not even beat a randomly initialized, untrained encoder, see Figure \ref{fig_A}. Across datasets, pretraining on 217 RDKit \cite{rdkit} descriptors is by far the most successful pretraining strategy for CLIMB.
We also trained models on experimental assay and transcriptomic labels drawn from the LargeMix collection~\cite{beaini2024foundational}---PCBA bioactivity and LINCS L1000 gene-expression profiles---datasets also used in training MiniMol, a successful GNN foundation model~\cite{klaser2024minimol}. These data are experimentally measured rather than computed and their coverage is highly incomplete (a median PCBA molecule carries 10 of 1,328 assay labels) and often sparse, where a few rare active molecules are crowded by many inactives. This ``sparse'' pretraining resulted in an exceptionally bad model, worse than a random encoder. Mixed in with dense descriptor data, performance increases, although not substantially so. Pretraining on unlabeled data followed by supervised pretraining did not improve downstream performance. Surprisingly, even the reverse, where we pretrained on labeled data first and then on unsupervised SMILES, while not improving on descriptor pretraining, did conserve its benefits.

Pure unsupervised training did lead to some performance gains over a random encoder. We included unsupervised pretraining in our further experiments to better understand what this model learns during pretraining. While the 8M  models generally have not yet plateaued in downstream performance,  they do retain their relative ranking. We were relieved to see that our two scaled-up models, pretrained on 100M molecules, perform competitively with three previously reported CLMs: MoLFormer \cite{ross2022molformer}, a large model pretrained on unlabeled data only; ChemBERTa-2 \cite{ahmad2022chemberta2}, a model pretrained on descriptors; and SELFIES-TED~\cite{selfies-ted}, a CLM using SELFIES \cite{selfies} instead of SMILES as input. We did also evaluate our best models in an end-to-end setting, where the loss from a downstream task propagates through the entire model, not just the prediction head, but found no improvements in performance, see SI section \ref{sec:end2end}.

\begin{figure}[t]
    \centering
    \includegraphics[width=\linewidth]{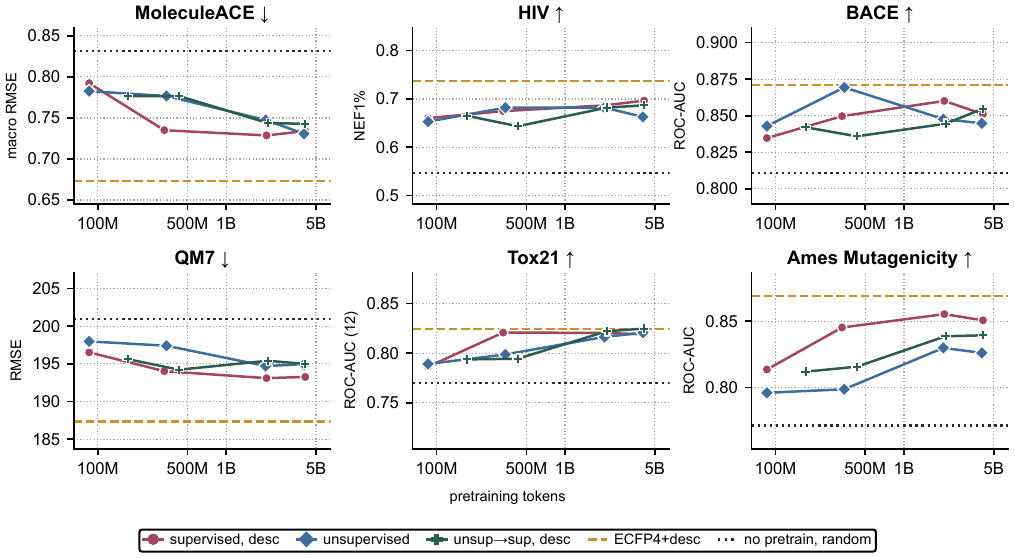}
    \caption{\textbf{Pretraining scaling.}
    Downstream performance as a function of pretraining tokens processed. Pretraining is enlarged by sampling more unique molecules from the corpus with all other settings staying constant.}
    \label{fig_B}
\end{figure}

However, the baseline XGBoost models dominated on virtually all datasets. These models, using the standard ECFP4 Morgan-fingerprint format \cite{morgan1965generation} in addition to the 217 RDKit descriptors \cite{rdkit}, have the best overall and also the most consistent performance across the 64 datasets evaluated. In many cases, the absolute gap in performance is striking, particularly on regression and activity cliff tasks, see SI Figure \ref{SI_fig_g_absolute_errors}. This cannot be said for XGBoost models using only fingerprints as input, which tend to be competitive on bioassay tasks like CBS or MoleculeACE while underperforming on datasets that heavily correlate with descriptors: ESOL and lipophilicity are two such tasks where performance quickly saturates using just a small number of descriptors. 

Even scaling up our pretraining efforts did not make our models competitive with this much simpler and faster baseline. As Figure \ref{fig_B} shows, scaling three CLIMB pretraining strategies---unsupervised, descriptor, and sequential pretraining---to 100M unique molecules while keeping model size fixed does not universally lead to performance gains on six representative tasks. While improvements occur, they tend to plateau early and at best approach the XGBoost baseline, as for Tox21. We find no robust and predictable downstream improvement with pretraining scale. This saturation could reflect limited model capacity or the fact that the additional molecules shown are irrelevant to downstream tasks. Regardless, our results suggest that the transferable information acquired during both supervised and unsupervised pretraining saturates relatively early and does not universally approach the performance of classical baselines.

\paragraph{Are Molecular and Task Similarity Relevant for Downstream Performance?}

\begin{figure}[t]
    \centering
    \includegraphics[width=\linewidth]{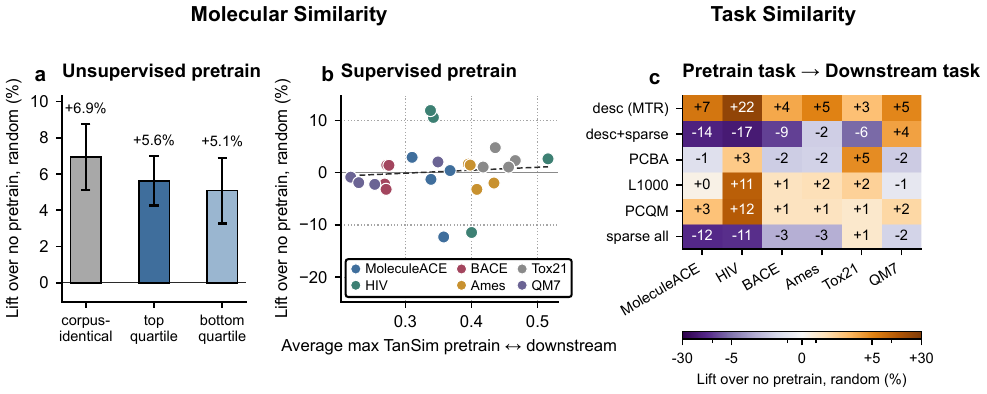}
    \caption{\textbf{Performance against pretraining and downstream similarity.}
    (a) Performance gain over a random encoder (``Lift'') for molecular structures that appear in the pretraining corpus (corpus-identical), as well as the top 25\% most similar and bottom 25\% least similar molecules. (b) Lift over a random encoder relative to the average nearest-neighbor distances across pretraining and downstream molecules.  (c) Lift over random for different pairs of pretraining datasets and downstream tasks, see SI Sections \ref{sec:downstream_datasets} and \ref{sec:SI_Methods} for further details on datasets.}
    \label{fig_C+D}
\end{figure}

Plausibly, downstream performance could be limited by the chemical space covered during pretraining. If so, models should perform better on molecules more similar to those seen during pretraining. We find little evidence for this, for either supervised or unsupervised pretraining (Figures \ref{fig_C+D}a,b). Performance is largely independent of corpus similarity, measured by fingerprint Tanimoto similarity between downstream molecules and the pretraining corpus; the gains observed do not appear significant relative to noise. Evaluating on more corpus-dissimilar molecules provides a soft proxy for out-of-domain generalization, where less similar molecules should be more challenging if the model learned specificities of its pretraining chemical space. 

Overall, similarity between pretraining and downstream tasks is not predictive of transfer. Reasonably, a model pretrained on labeled data could be expected to perform better on similar downstream tasks. We find no such evidence: Surprisingly, when comparing descriptor to bioassay pretraining, descriptor-based pretraining wins on both smooth, descriptor-like tasks as well as on bioassay-like downstream datasets with the gap in performance actually widening on the latter. 
Figure \ref{fig_C+D}c shows that descriptor pretraining is consistently helpful, while other transfer patterns are less intuitive. For example, PCBA, a bioassay classification task, transfers best to the toxicity task Tox21, while PCQM, a quantum-mechanical regression task, transfers best to the sparse HIV virtual screen. We interpret this as descriptor pretraining teaching the model real, physico-chemical information during pretraining. Plausibly, the smooth distribution of descriptor labels might be a more useful training signal compared to the sparse, incomplete bioassay datasets. We test this by pretraining on a version of the descriptor corpus, where descriptors are randomly shuffled across molecules.

\paragraph{Does Pretraining Require Real Chemistry?}

\begin{figure}[t]
    \centering
    \includegraphics[width=\linewidth]{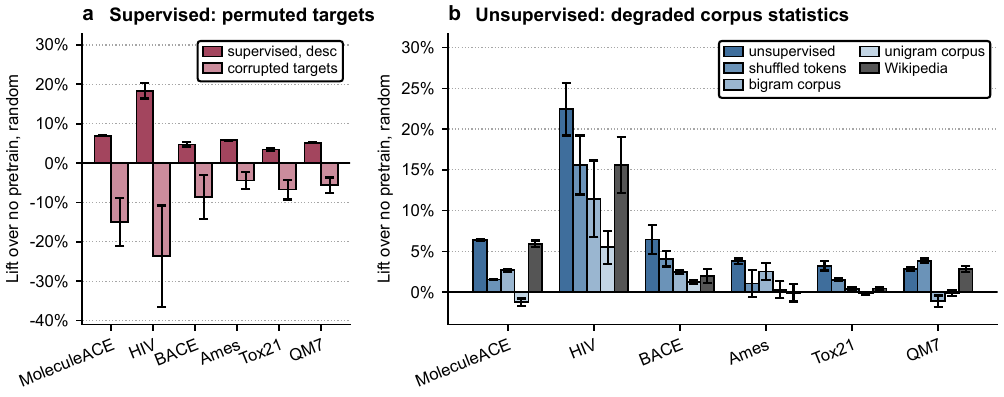}
    \caption{\textbf{Corrupted pretraining objectives.}
    Downstream lift defined as performance gains over a random encoder after corrupting supervised molecule-descriptor mappings (a) or progressively removing chemical structure from the unsupervised corpus (b).}
    \label{fig_E}
\end{figure}

Ideally, the pretraining of CLMs teaches generalizable, chemical concepts that can be usefully transferred to downstream tasks. To test this, we first used a corrupted pretraining target on our model pretrained on descriptor data, where we swapped descriptor values at random across the dataset, see Figure \ref{fig_E}a. Crucially, this does not change the marginal distribution of any descriptor but rather removes the linkage between molecular structure and its series of physico-chemical properties. This indeed led to a catastrophic loss of performance, where the model now underperforms even a random encoder. Seemingly, the model does learn a mapping between the structure of a molecule and its descriptors that goes beyond simply memorizing the distribution of labels. Indeed, the descriptors of a molecule can be predicted using a simple ridge head from the embeddings of the descriptor-pretrained model with a median $R^2$ of 0.82, whereas both the unsupervised (0.59) and random (0.34) variants perform substantially worse, see SI Table \ref{tab:descriptor_probe}. We consider this convincing evidence that supervised pretraining does reorganize a CLM's latent space along exactly these physico-chemical properties. 

A very different story emerges when corrupting the unsupervised pretraining corpus. \textit{A priori}, only filling in missing tokens in a SMILES string should not teach a model much chemistry beyond some low-level statistics of the corpus itself. Here, we use a series of perturbations to figure out what low-level statistics the model might be learning during pretraining. Firstly, we shuffle the tokens within a SMILES string while maintaining the co-occurrence of these tokens within a molecule. We then use bigram sampling, which keeps the frequency of local adjacencies of tokens intact, but alters which tokens occur in a molecule, as well as unigram sampling, where we sample tokens from the corpus marginal, i.e., the individual token frequencies. As an example of an entirely non-chemical corpus, we also tokenize English natural-language Wikipedia articles \cite{wikidump} using the same tokenizer we use for all CLIMB models. All methods, including the Wikipedia sampling, maintain the sequence length distribution of SMILES relative to the original corpus. We were astonished to see how much of the benefit of unsupervised pretraining can be recovered by pretraining on Wikipedia texts, particularly on the MoleculeACE and HIV tasks. 

We find evidence for three separable mechanisms at play during unsupervised pretraining, see Figure \ref{fig_E}. Firstly, unsupervised pretraining leads to a better initialized model. We isolate the effect of better initialized weights after unsupervised pretraining by sampling the mean and variance of weights from a pretrained model without training on any data, see SI Section \ref{sec:SI_weights_conditioning}. We find that better initialization accounts for between 8--28\% (and 100\% in the case of QM7) of the gain seen in unsupervised pretraining. Secondly, we do find evidence that our model learns low-level statistics on token adjacencies and per-molecule co-occurrences as both the bigram and shuffled-tokens experiments lead to some improvement over just weight conditioning, particularly on HIV and MoleculeACE. Lastly, we find that some results only arise when training on real SMILES, particularly BACE and Tox21, where the actual order of tokens in a SMILES seems to matter as well. In summary, we find that unsupervised pretraining acts both through improved initialization of weights as well as through learning low-level statistics on the pretraining corpus, although these mechanisms do not necessarily require real chemical data.

\paragraph{Do CLM Embeddings Supply Useful Information Beyond Classical Features?}

\begin{figure}[t]
    \centering
    \includegraphics[width=\linewidth]{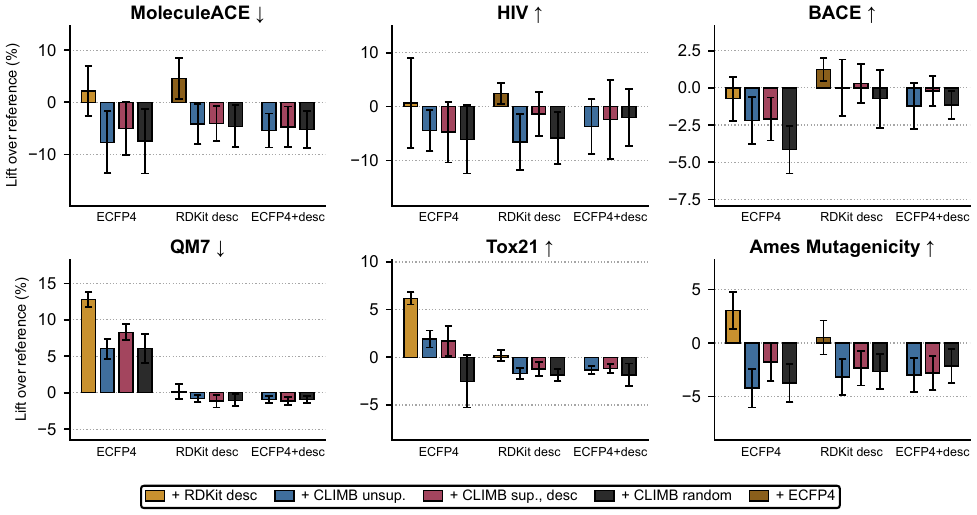}
    \caption{\textbf{What information do CLIMB embeddings carry?} CLIMB embeddings were appended to either fingerprints, descriptors, or both and combinations were used as input to an XGBoost model to isolate how much additional information our embeddings carry. We report the improvements in prediction performance, downstream lift, relative to the respective feature set named on the x-axis.}
    \label{fig_F}
\end{figure}

To see whether our CLIMB embeddings carry any supplementary information relative to fingerprints and descriptors, we tested what concatenating our embeddings adds in performance relative to using fingerprints, descriptors, or the combination of both. All combinations are fed into the same XGBoost prediction head with five-fold cross-validation. Figure \ref{fig_F} shows that CLIMB embeddings are generally not very helpful for these downstream tasks. In no case does adding the CLIMB embeddings improve performance over fingerprints+descriptors, suggesting that CLIMB embeddings carry no useful information over fingerprints and descriptors. They do seem to add some value relative to using the fingerprint in isolation, at least for the Tox21 and QM7 tasks. Whereas MoleculeACE, HIV, and BACE all can be well predicted with fingerprints alone, even descriptors adding little value here, the other three remaining tasks benefit from adding descriptors substantially. This is consistent with the view that CLIMB embeddings mostly supply physico-chemical information when pretrained on descriptors in addition to being good structure encoders by default.

Alternatively, these embeddings might simply help by providing a smooth molecular representation, contrary to the sparse, discrete ECFPs, which are useful for predicting smooth tasks like Tox21 and QM7. At least on the QM7 task, the random encoder is as useful as the unsupervised CLIMB model. Why this would be the case for one classification task, Tox21, but not another, BACE, is unclear however. Overall, we find that CLIMB embeddings carry no useful additional information for molecular property prediction. We do find that the supervised CLIMB embeddings are better than the unsupervised embedding as well as ECFP at predicting descriptors, see Section \ref{sec:reconstructing_desc_from_emb} in the SI. Evidently, they do carry some information on the physico-chemical properties of a molecule beyond what is in the fingerprint alone.

\paragraph{What Molecular Changes Can CLM Embeddings Resolve?}

\begin{figure}[t]
    \centering
    \includegraphics[width=\linewidth]{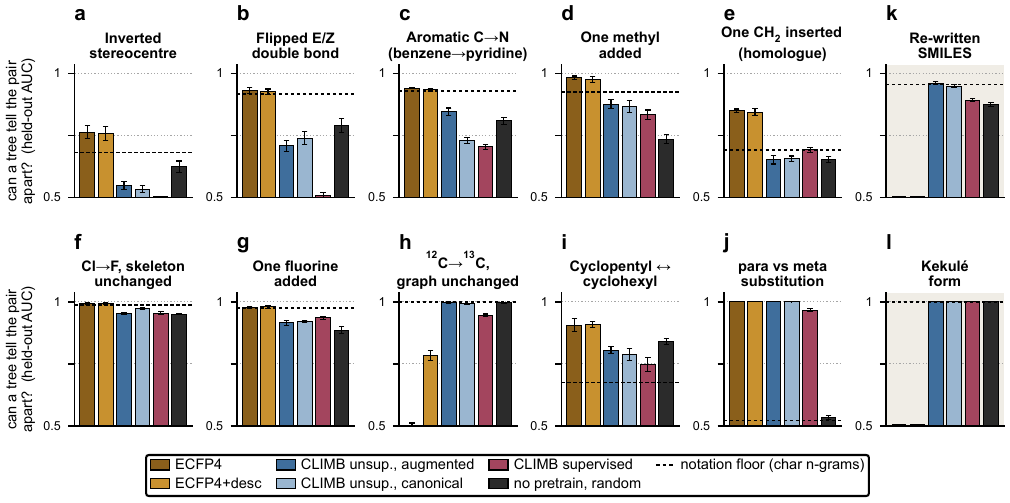}
    \caption{\textbf{Can representations resolve chemical edits?} We tested ten chemical edits (1{,}000 molecule pairs each), where each parent is labeled 0 and every edited partner 1, giving 2{,}000 labeled vectors. An XGBoost model is tasked with predicting the labels. We report the area under the curve (AUC) of the receiver operating characteristic, where 0.5 would correspond to random predictions. Notation-only edits (SMILES enumeration, Kekulization) should score $0.5$.}    
    \label{fig_G}
\end{figure}

Ultimately, to be successful at any molecular property prediction task, the embeddings used as input must resolve the differences in chemical structure relevant to a prediction. For example, inverting a stereocenter  can destroy the activity of a drug as biology is homochiral. Similarly, activity cliffs are often dependent on adding a single functional group or a single-atom scaffold edit. If these changes are collapsed into indistinguishable embeddings already at the encoding step, no prediction head will ever successfully separate these molecules. We hence analyzed whether XGBoost models, which split data on at least one feature dimension, can resolve molecules that differ in only one isolated chemical edit. We report the Area Under the Curve (AUC) of the Receiver Operating Characteristic for the trained models, where 0.5 is equivalent to a random model, as well as an n-gram baseline, which simply counts  SMILES characters and character pairs. We find that  fingerprints alone already resolve nearly all cases, including stereochemistry, single-atom changes, and structural homologues, see Figure \ref{fig_G}. Adding descriptors resolves the one change fingerprints alone cannot detect: isotope substitution. The discrete nature of such fingerprints allows these molecules to be split along often only a few altered dimensions, lending itself naturally to the inductive bias of a model such as XGBoost \cite{chen2016xgboost, xia2023understandingXgboost}. 

We also find that CLIMB models generally do resolve most chemical edits with the exception of stereochemistry. However, they do so no better than a randomly initialized, untrained CLIMB model. By a Johnson--Lindenstrauss-type argument \cite{johnson1984extensions}, the untrained transformer block approximately acts as a random linear projection of the token count vector, which produces consistent displacements across different molecules. In fact, the random encoder manages to resolve stereochemistry, which involves a token-level change from \texttt{@} to \texttt{@@}, which the pretrained CLIMB models learn to ignore during the stereo-invariant pretraining. As expected, the random encoder fails to resolve molecules that have the same token set, as between a para- versus meta-substituted molecule. Similarly, pretraining on descriptors prevents the CLIMB model from resolving double-bond stereochemistry, which also does not change the physico-chemical properties of a molecule. All CLIMB models are capable of resolving isotope substitution of $^{12}$C to $^{13}$C, which changes one token explicitly from \texttt{C} to \texttt{[13C]}. This same sensitivity becomes a disadvantage when considering the same molecule written as different SMILES. Here, ECFP4 correctly uses identical embeddings, while different SMILES produce very different embeddings in the case of CLIMB models. This failure mode is not meaningfully reduced when pretraining on augmented SMILES. In this case, CLIMB embeddings struggle to correctly encode molecular invariances. 

Surprisingly, pretraining seems to make models no better at encoding molecular structure in a way that a prediction head like XGBoost finds useful. These results suggest that CLMs encode the same amount of structural information as classical molecular features with the advantage of using fewer features (512 rather than 2{,}048 or 2{,}265 dimensions). The disadvantage of such a compressed embedding is that chemical edits are distributed across many dimensions, which can limit the effectiveness of tree-based prediction heads such as XGBoost \cite{xia2023understandingXgboost}. We find that our pretrained CLMs resolve molecular structure no better than a random encoder and slightly worse than a fingerprint- and descriptor-based embedding. 

\section{Discussion and Conclusion}

We find that pretraining can improve downstream performance in CLMs. However, the worst pretraining settings underperform even a random encoding model and we find that the benefits of pretraining are quickly saturated and do not scale with more data. Only supervised pretraining on physico-chemical descriptors reliably delivered good downstream performance. We control for some possible confounding factors: (i.) we find no evidence that training our CLIMB models end-to-end on a downstream dataset improves performance (SI Figure \ref{SI_fig_a_end2end_vs_pretrain}); (ii.) using XGBoost rather than MLP prediction heads for CLIMB models decreases performance slightly (SI Figure \ref{SI_fig_f_predictionHead}), throughout the paper we report the MLP-based results; (iii.) we do not find that our models are bottlenecked by our specific choice of tokenizer or vocabulary size (SI Figure \ref{SI_fig_b_tokenizer}); (iv.) we used unsupervised pretraining with both canonical and augmented SMILES and found no difference in downstream performance (SI Figure \ref{SI_fig_d_canon_vs_enum}) nor are these models meaningfully better at recognizing degenerate SMILES, see Figure \ref{fig_G}. Lastly, we also find no evidence for the often-repeated claim that pretrained models are better in a low-data regime. In fact, across baseline, pretrained, and end-to-end fine-tuned models the gaps in performance remain surprisingly static across dataset size, see SI Figure \ref{SI_fig_e_end2end_overtake}. 

It seems that supervised pretraining is most successful precisely because it teaches physico-chemical properties. The CLIMB embeddings resulting from supervised descriptor pretraining seem to contain similar information content to descriptors, see Figure \ref{fig_F}, and indeed descriptors can be easily reconstructed from these embeddings, see SI Table \ref{tab:descriptor_probe}. The evidence in the case of unsupervised pretraining seems to hint at models learning low-level statistics that can be useful in some prediction tasks. However, these improvements can even be observed when training on non-chemical corpora, in our case English-language Wikipedia articles. While models unambiguously improve over an untrained encoder, we remain unconvinced that much real chemistry is being learned. We find no evidence that CLMs learn any chemical intuition beyond structural representation and physico-chemical properties that, alternatively, could also have been computed exactly. Under all settings evaluated, classical encodings outperform pretrained models while obviating the need for computationally intensive pretraining. 

\paragraph{Limitations} Our conclusions are restricted to the 41.4M-parameter, encoder-only CLIMB architecture and the supervised and unsupervised objectives studied here. Our conclusions may not extend to other CLM architectures, such as decoder-only or encoder--decoder transformers, and do not necessarily generalize to different architectures such as GNNs. It is conceivable that the true benefits of pretraining may only arise when using much larger models, substantially larger datasets, alternative objectives, or using very different architectures. In particular, we do not consider reaction prediction as a pretraining task \cite{Schwaller2019} nor domain-specific objectives \cite{spence2025smilest5domainspecificpretrainingmolecular, fabian2020molbert}.  

\section*{Acknowledgments}

We thank Amazon Web Services (AWS) and Y Combinator (YC) for generous compute credits used for model training and evaluation.

\section*{Code and Data Availability}

All code for this paper is available at \url{https://github.com/leifsieben/CLIMB}.  Pretrained encoder weights for all models are available at
\url{https://huggingface.co/lsieben/climb-encoders}. Full evaluation results are available at \url{https://huggingface.co/datasets/lsieben/climb-results}. The pretraining corpora are available
at \url{https://huggingface.co/datasets/lsieben/climb-pretrain-data}.

\bibliographystyle{unsrtnat}
\bibliography{references}  


\appendix
\newpage
\section*{Supplementary Information}

\renewcommand{\thefigure}{\thesection.\arabic{figure}}  
\renewcommand{\thetable}{\thesection.\arabic{table}}
\renewcommand{\theHfigure}{\thesection.\arabic{figure}}
\renewcommand{\theHtable}{\thesection.\arabic{table}}
\counterwithin{figure}{section}   
\counterwithin{table}{section}

Here, in addition to an extended methods section, we report a number of results on potentially confounding factors regarding our downstream evaluation such as the choice of prediction head, tokenizer and vocabulary size, and use of augmented or canonical SMILES during pretraining. We also test whether end-to-end training, from either a pretrained checkpoint or from scratch, can yield better downstream performance. 

\section{Methods}
\label{sec:SI_Methods}

\subsection{Model architecture} All CLIMB encoders follow the same architecture: a 12-layer ModernBERT-style bidirectional transformer with 41.4M parameters. Alternating local and global attention (a 128-token sliding window, with full attention every third
layer) keeps the cost linear in sequence length while preserving long-range structure. The tokenizer uses byte-pair encoding (BPE) with a 1{,}000-token vocabulary fit on a 10M-molecule sample of the filtered PubChem corpus, and it is held fixed across all runs in the paper, including those pretrained on the larger corpus.

The only differences between the models compared in this paper are the {pretraining objective}, the {pretraining corpus} and the {compute budget} with all other settings held constant. We refer to models based  on the number of unique molecules seen during pretraining (either 8M or 100M). Pretraining runs are measured in individual tokens. Downstream, the encoder is frozen and only a small probe head is trained (Section~\ref{sec:downstream_datasets}).

\begin{table}[tbh]
\centering
\caption{\textbf{Encoder architecture and optimizer settings.}}
\label{tab:arch}
\small
\begin{tabular}{@{}ll@{}}
\toprule
Component & Setting \\
\midrule
Backbone & ModernBERT (RoPE, pre-norm, GeGLU, global/local attention) \\
Layers / hidden size / heads & 12 / 512 / 8 \\
Feed-forward width (GeGLU) & 1{,}536 \\
Global attention & every 3 layers; local window 128 \\
Position encoding & rotary; max position 256 \\
Dropout (attn / mlp / embed) & 0 / 0 / 0 \\
Parameters & $\approx$41.4\,M \\
Tokenizer & byte-level BPE, vocabulary 1{,}000 (fixed) \\
Sequence length (train / eval) & 128 / 256 \\
\midrule
Optimizer & AdamW~\cite{loshchilov2019adamw} \\
Learning rate / warm-up & $2\times10^{-4}$ / 5\% ratio \\
Weight decay / grad-clip & 0.01 / 1.0 \\
Batch size / precision & 256 / bfloat16 \\
MLM masking rate & 30\% selection rate (80\% \texttt{[MASK]} / 10\% random / 10\% kept) \\
\bottomrule
\end{tabular}
\end{table}

\subsection{Pretraining datasets} CLIMB is pretrained using three types of objective: The unsupervised objective is masked language modeling on SMILES alone. The descriptor objective (multi-task regression, MTR) regresses 217 precomputed RDKit \cite{rdkit} descriptors. These are deterministic functions of structure rather than measured labels. The {assay}
objective is multi-task supervised learning on real measured labels drawn from four public panels,
following the MiniMol LargeMix selection \cite{klaser2024minimol, beaini2024foundational}. In certain pretraining settings, multiple objectives are run sequentially or interleaved where data of two sources is combined and the objective is chosen based on which training data is sampled at each step. 

Two training corpora are used. Most models train on a filtered PubChem set of $\approx$12M molecules \cite{kim2025pubchem, heiden2026pubchem124m}. The large-scale runs train on the full 124M-molecule PubChem release. Descriptor targets are $z$-scored using statistics fit once on the corpus and subsequently held fixed. Assay datasets are subsampled with per-family caps such that no single family dominates the mixture.

\begin{table}[tbh]
\centering
\caption{\textbf{Pretraining data.}}
\label{tab:pretrain_data}
\small
\begin{tabular}{@{}llrrl@{}}
\toprule
Source & Objective & Labels & Molecules & Label distribution \\
\midrule
\multicolumn{5}{@{}l}{\emph{Unsupervised corpora}} \\
PubChem, filtered & MLM, MTR & --- &  12,000,000 & --- \\
PubChem, full & MLM, MTR & --- & 123{,}400{,}000 & ---\\
\midrule
\multicolumn{5}{@{}l}{\emph{Computed targets}} \\
RDKit descriptors & MTR & 217 & all of the above & continuous, $z$-scored, clipped $\pm10$ \\
\midrule
\multicolumn{5}{@{}l}{\emph{Assay label families}} \\
PCQM4M & supervised & 25 & 3{,}810{,}323 & continuous \\
PCBA & supervised & 1{,}328 & 1{,}563{,}775 & binary; $10.8\%$ of cells measured \\
L1000 MCF7 & supervised & 978 & 11,429 & continuous, $z$-scored expression \\
L1000 VCAP & supervised & 978 & 14,953 & continuous, $z$-scored expression \\
\bottomrule
\end{tabular}

\vspace{3pt}
\end{table}

\subsection{Downstream benchmarking datasets} Every model is evaluated using 64 datasets grouped into four task types (activity cliffs, virtual screening, classification, regression): five MoleculeNet tasks \citep{Wu2018MoleculeNet} (BACE, ESOL, HIV, QM7 and Tox21), 30 ChEMBL potency-regression targets from MoleculeACE \citep{van2022Exposing}, 26 Polaris benchmarks \citep{wognum2024polaris} comprising 17 ADMET and toxicity tasks from the Therapeutics Data Commons
\citep{huang2021tdc_polaris}, six Fang ADME assays \citep{fang2023adme_polaris} and
three PKIS2 kinase regressions \citep{drewry2017pkis2_polaris}, and three
single-dataset benchmarks: a cystathionine $\beta$-synthase virtual screen
\citep{Truong2026Enhancing}, a \textit{Staphylococcus aureus} antibacterial screen by
\citet{wong2024discovery}, and the multi-class taste dataset by
\citet{zimmermann2025taste}. We use a representative subset of six datasets to report task-level results covering potency regression under activity cliffs, a rare-active virtual screen, a homogeneous single-series binary task, mutagenicity, a 12-assay toxicity panel, and a quantum-chemical regression, see Table~\ref{tab:downstream_data}.  The encoder is
frozen throughout and only the probe head is trained, so the comparison isolates the effect of molecular representations.

\label{sec:downstream_datasets}

\begin{table}[tbh]
\centering
\caption{\textbf{Panel of downstream datasets.} }
\label{tab:downstream_data}
\small
\begin{tabular}{@{}lllrr@{}}
\toprule
Dataset & Task type & Metric & Molecules & Actives \\
\midrule
MoleculeACE & potency regression, 30 targets & macro RMSE & 48{,}714 & --- \\
HIV & rare-active screen & NEF1\% & 41{,}120 & 1{,}443 (3.51\%) \\
BACE & binary classification & ROC-AUC & 1{,}513 & 691 (45.7\%) \\
Tox21 & binary classification, 12 assays & mean ROC-AUC & 7{,}823 & 5{,}858 (7.52\%) \\
QM7 & quantum regression & RMSE (kcal/mol) & 6{,}838 & --- \\
Ames & binary classification & ROC-AUC & 7{,}278 & $\approx$53.3\%\textsuperscript{a} \\
\bottomrule
\end{tabular}
\vspace{3pt}

{\footnotesize $^{\textrm{a}}$Polaris withholds the test labels, so the positive rate is measured on the train split.}
\end{table}

\subsection{Error bars}  Normally, a 5-fold Bemis--Murcko scaffold cross-validation is run, averaging the three prediction head seeds within each fold, where each is scored once per fold. The reported interval is the spread across the five folds. As Polaris withholds its test labels, the reported interval is instead the spread across the three replicate seeds for each split.

\section{Resolving power of embeddings}

Figure \ref{fig_G} attempts to quantify how well different embeddings can distinguish molecules that are globally similar but differ in one chemical edit (e.g. stereoinversion). We first curated a pool of unique, RDKit-canonicalized molecules from the benchmark datasets (30 MoleculeACE ChEMBL targets, BACE and Tox21). From this we generated 1,000 matched molecule pairs for each of ten structural edits, together with two notation-only controls (SMILES re-enumeration and Kekulization) in which the string changes but the molecule does not. Structural edits were compared as canonical SMILES, so that the only string difference is the one the edit forces. For each representation and each edit, every parent molecule was labeled 0 and its edited partner 1, giving 2{,}000 labeled vectors, and a gradient-boosted tree classifier (XGBoost; 200 trees, maximum depth 4, learning rate 0.3) was trained to separate the classes. Because molecules recur across pairs, pairs sharing a molecule were merged into connected components by union such that no test molecule appears in training. 20\% of pairs were held out, and we report the mean held-out ROC-AUC over five split seeds ($\pm$ 1 s.d.). As a baseline, we use the character 1- and 2-grams of the representative SMILES string and also pass it through the XGBoost model.

\section{Absolute errors on benchmarking tasks}

\begin{figure}[tbh]
    \centering
    \includegraphics[width=\linewidth]{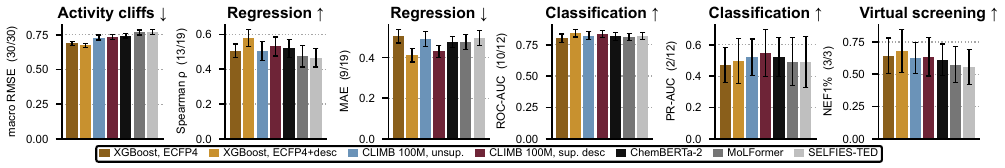}
    \caption{\textbf{Absolute errors on benchmarking suite.}
    To show the absolute performance of various models rather than just relative rankings, we also report the average error per dataset type.}
    \label{SI_fig_g_absolute_errors}
\end{figure}

To better understand the absolute performance of our models rather than just reporting relative rankings as in Figure \ref{fig_A}, we display the absolute errors across the four main categories of tasks these models were evaluated on, see Figure \ref{SI_fig_g_absolute_errors}. We find that the best XGBoost model is particularly dominant on activity cliffs and regression, while performance differences are not distinguishable from noise for classification and virtual screening. Nonetheless, XGBoost on fingerprints+descriptors is by far the most competitive, robust, and generalizable prediction method across deployment settings examined. 

\section{Reconstruction of descriptors from CLIMB embeddings} 
\label{sec:reconstructing_desc_from_emb}

To evaluate if supervised pretraining on descriptors actually results in the CLIMB model learning to reorder its latent space according to these physico-chemical properties we tried predicting these descriptors from the embeddings. We compare this to predicting descriptors from ECFPs, the CLIMB unsupervised embeddings, and a randomly initialized CLIMB model. We use a ridge regression head on all embeddings.

\begin{table}[h!]
\centering
\footnotesize
\setlength{\tabcolsep}{4.5pt}
\caption{\textbf{Descriptor prediction from embeddings.} Median $R^2$ of a
ridge head fit from the frozen representation to each of 217 RDKit descriptors, on 1,940
held-out molecules. Targets are $z$-scored with the pretraining corpus statistics and clipped at $\pm10$, matching the descriptor objective.}
\label{tab:descriptor_probe}
\begin{tabular}{@{}l cccc cccc @{\hspace{6pt}}||@{\hspace{6pt}} c@{}}
\toprule
& \multicolumn{4}{c}{\emph{topology and shape}}
& \multicolumn{4}{c}{\emph{substructure and electronic}} & \\
\cmidrule(lr){2-5} \cmidrule(lr){6-9}
\textbf{Representation} & \textbf{Chi} & \textbf{Kappa} & \textbf{BCUT} & \textbf{Ipc/SPS} & \textbf{EState} & \textbf{VSA} & \textbf{fr\_*} & \textbf{Bulk} & \textbf{ALL} \\
\midrule
ECFP4 (2{,}048-bit) & 0.73 & 0.69 & 0.66 & 0.69 & 0.65 & 0.69 & 0.61 & 0.72 & 0.69 \\
CLIMB unsup & 0.86 & 0.79 & 0.73 & 0.75 & 0.56 & 0.62 & 0.33 & 0.70 & 0.59 \\
CLIMB sup, desc & \textbf{0.97} & \textbf{0.95} & \textbf{0.86} & \textbf{0.89} &\textbf{ 0.87} & \textbf{0.77} & \textbf{0.74} & \textbf{0.91} & \textbf{0.82} \\
CLIMB random & 0.42 & 0.34 & 0.53 & 0.42 & 0.31 & 0.34 & 0.16 & 0.41 & 0.34 \\
\bottomrule
\end{tabular}
\vspace{2pt}
\end{table}

Overall, these results confirmed our underlying hypothesis that a CLIMB model is able to learn descriptor chemistry, its median $R^2$ value is by far the highest and actually beats the other embeddings on every individual descriptor recorded, see Table \ref{tab:descriptor_probe}. Overall, the fingerprint embedding does beat the unsupervised CLIMB embedding, although the unsupervised CLIMB embeddings are markedly better at recovering topological and shape descriptors. The unsupervised embeddings are also substantially better at predicting descriptors than a randomly initialized model, suggesting that even unsupervised pretraining CLIMB embeddings do retain some meaningful molecular information, presumably the ability to cluster similar structures closer together. 

\section{Deploying pretrained models end-to-end does not improve downstream performance}
\label{sec:end2end}

\begin{figure}[h]
    \centering
    \includegraphics[width=\linewidth]{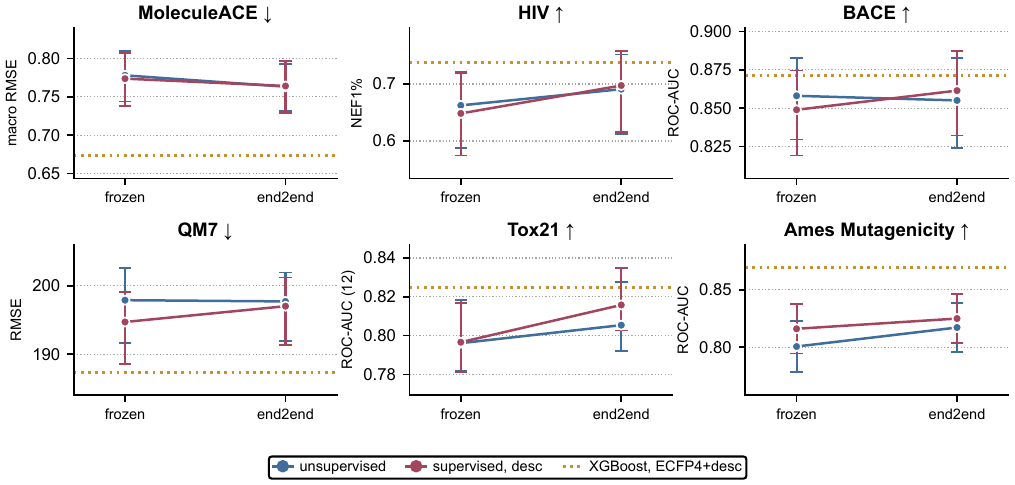}
    \caption{\textbf{Is end-to-end fine-tuning necessary?}
    The same pretrained encoder used two ways at full downstream data: frozen with a trained probe,
    versus the whole network fine-tuned. Intervals are 95\% resampling intervals. Dotted line: XGBoost on ECFP4+descriptors. }
    \label{SI_fig_a_end2end_vs_pretrain}
\end{figure}

We mostly evaluated our CLIMB models in a frozen encoder setting, where we only train the prediction head on the downstream task. To see if this is also a major bottleneck for our CLIMB models we chose the best supervised model trained on descriptors and the model pretrained on unlabeled data only and compared their performance in both a frozen encoder and end-to-end setting. We find that training end-to-end, which involves significantly more compute, can have positive effects in some larger datasets, namely the HIV virtual screening and the Tox21 dataset, but overall only equates to a marginal advantage, never beating the best XGBoost baseline methods. End-to-end is ahead in 12 of 18 cells tested, but differences are not meaningfully resolved at this test set size, we report six of these in Figure \ref{SI_fig_a_end2end_vs_pretrain}. The CLIMB models do not substantially benefit from end-to-end fine-tuning. 

\section{Downstream performance is not bottlenecked by the prediction head}

\begin{figure}[h]
    \centering
    \includegraphics[width=\linewidth]{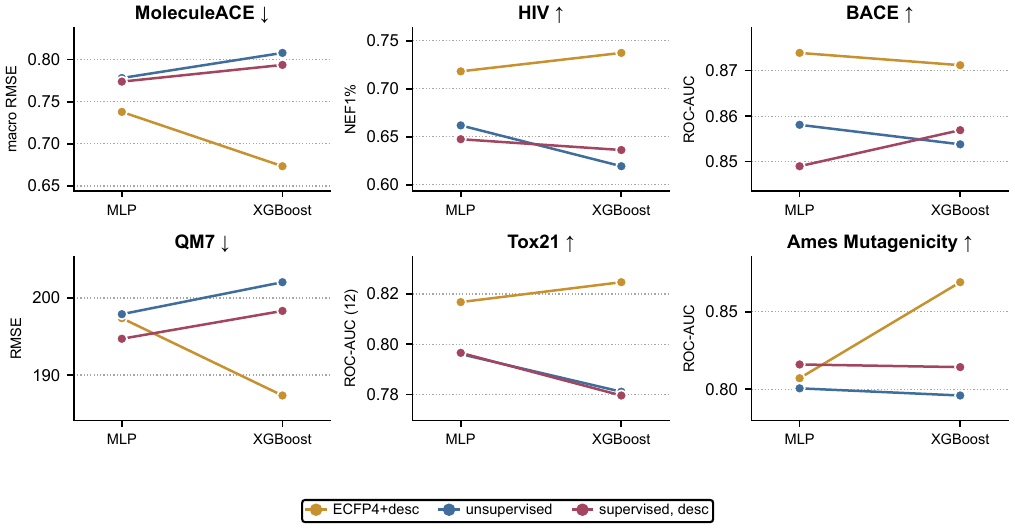}
    \caption{\textbf{The probe head is not neutral.}
    The same frozen representations scored through an MLP head and through XGBoost, everything else
    fixed. The preference is representation-dependent and consistent across panels: the CLIMB
    encoders do better with the MLP head, while fingerprints and descriptors  do better
    with XGBoost. In the main text, each representation is reported at its own best head in Fig.~1.}
    \label{SI_fig_f_predictionHead}
\end{figure}

To control for the use of prediction head, we also evaluated two of our CLIMB models (supervised, descriptor and unsupervised)  using both their default MLP head as well as XGBoost, see Figure \ref{SI_fig_f_predictionHead}. We find no systematic benefit in using XGBoost on CLIMB embeddings and the MLP is the better choice on average. This makes some intuitive sense as XGBoost splits the dataset on specific features which is easier with discrete, mostly orthogonal dimensions in a fingerprint rather than a continuous embedding where molecular changes are mixed across many dimensions. It speaks to the robustness of XGBoost that it is able to remain competitive at all with an MLP head even using CLIMB embeddings. 

\section{Downstream performance is independent of the tokenizer and vocabulary size used}

\begin{figure}[h!]
    \centering
    \includegraphics[width=\linewidth]{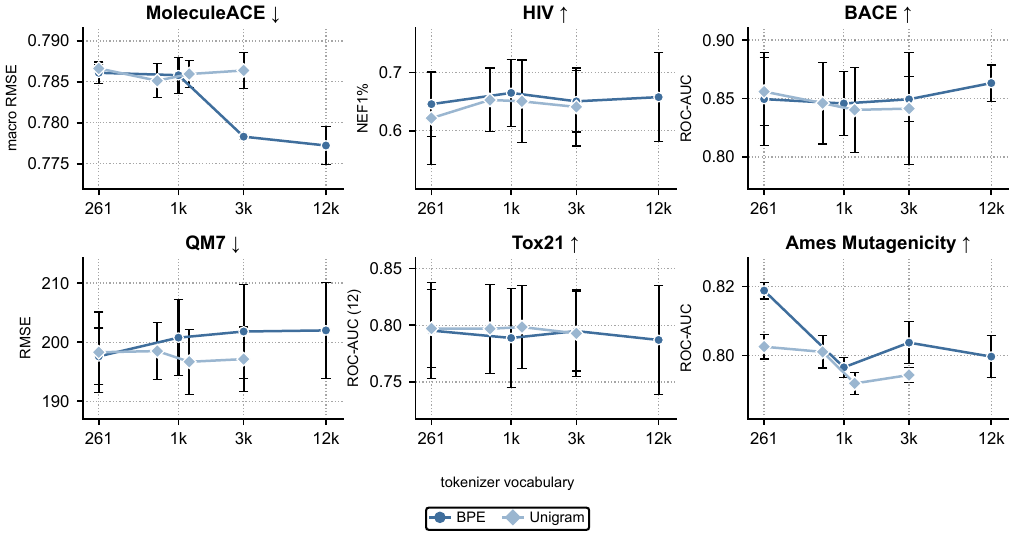}
    \caption{\textbf{Tokenizer family and vocabulary size.} We compare two tokenizer types: byte-level BPE and Unigram-LM. The actual vocabulary sizes achieved on the pretraining corpus are: BPE $\{261, 1000, 3000, 12000\}$; Unigram $\{261, 700, 876, 3000\}$. No discernible effect on downstream performance is found.}
    \label{SI_fig_b_tokenizer}
\end{figure}

We keep CLIMB's underlying architecture constant across all experiments unless otherwise stated. One potential confounding factor we were interested in was the vocabulary size and choice of tokenizer. No general consensus exists on the optimal architecture for CLMs and a number of different tokenizers have been used \cite{sultan2024review}. The two tokenizers work in opposite directions: BPE starts with basic characters, finding the most common pairs and merging them until the target vocabulary size is reached; Unigram-LM starts from all possible subwords, computes token probabilities, dropping tokens that lower the loss least until the target vocabulary size is reached. 

Both of these tokenizers need to be briefly pretrained themselves to learn the tokens on a representative corpus. We pretrain both tokenizers on the same corpus and aim for similar vocabulary sizes, although the exact vocabulary reached varies across them. We then use the resulting tokenizers in the same unsupervised-only pretrained CLIMB model which we evaluate on the six tasks shown in Figure \ref{SI_fig_b_tokenizer}. Reassuringly we find that CLIMB's performance is essentially independent of both tokenizer and vocabulary size choice. If anything, larger vocabulary sizes seem to hurt performance for the BPE tokenizer although  most differences are well within the noise of the five cross-validation folds. This largely agrees with prior work finding that tokenization has only a limited effect on downstream property prediction~\cite{Wadell2026Tokenization,heidenreich2026cutdeepbpeunigramlm}.

\section{Downstream performance as a function of downstream dataset size}

We had seen some evidence that fine-tuning a randomly initialized CLIMB model only on downstream data, i.e. no pretraining, can be competitive on large datasets. We were thus curious to test whether there exists a clear intersection point in terms of dataset size where training end-to-end becomes competitive with a pretrained CLIMB model. By downsampling the same dataset, we benchmark the supervised, descriptor and unsupervised-only CLIMB model against the end-to-end versions of CLIMB. We find that if a gap in performance exists, it is essentially stable across dataset size and that our initial hypothesis that end-to-end training without pretraining becomes increasingly competitive as downstream datasets grow in size could not be confirmed. 

\begin{figure}[tbh!]
    \centering
    \includegraphics[width=\linewidth]{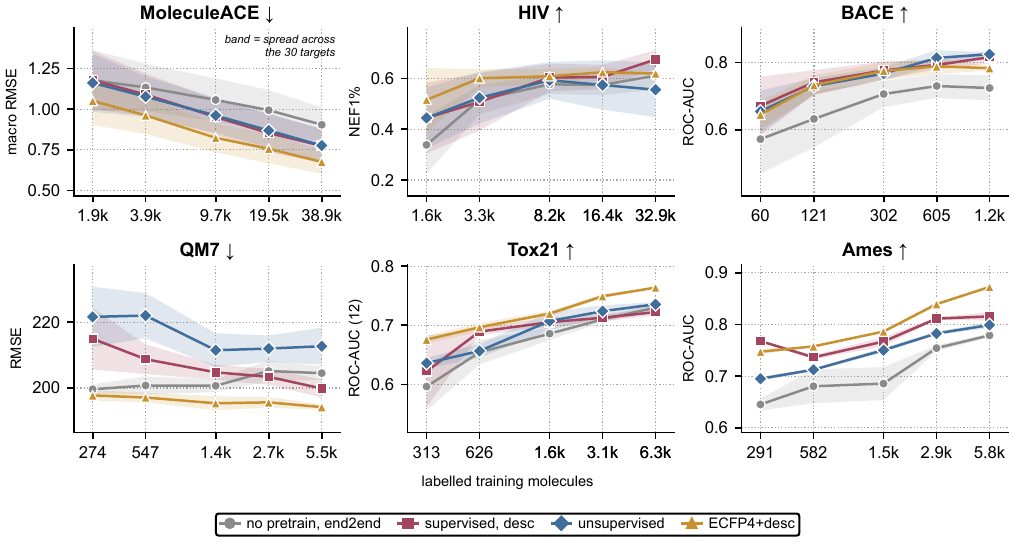}
    \caption{\textbf{Label efficiency.}
    Absolute performance against the number of labeled training molecules (log scale), with an
    identical hold-out split, label fractions and seed grid across models.}
    \label{SI_fig_e_end2end_overtake}
\end{figure}

\section{Using canonical versus augmented SMILES in unsupervised pretraining does not impact downstream performance}

The use of SMILES enumeration for data augmentation is an established technique in CLMs \cite{bjerrum2017smilesenumerationdataaugmentation}. Nonetheless, we were curious to test whether enumeration generalizes better than exact repetition in the multi-epoch regime. We thus compared presenting each molecule in its exact canonical form (repetition) against fresh on-the-fly RDKit SMILES enumeration while doing multiple epochs over a $\approx$12\,M-molecule corpus. 

\begin{figure}[t]
    \centering
    \includegraphics[width=\linewidth]{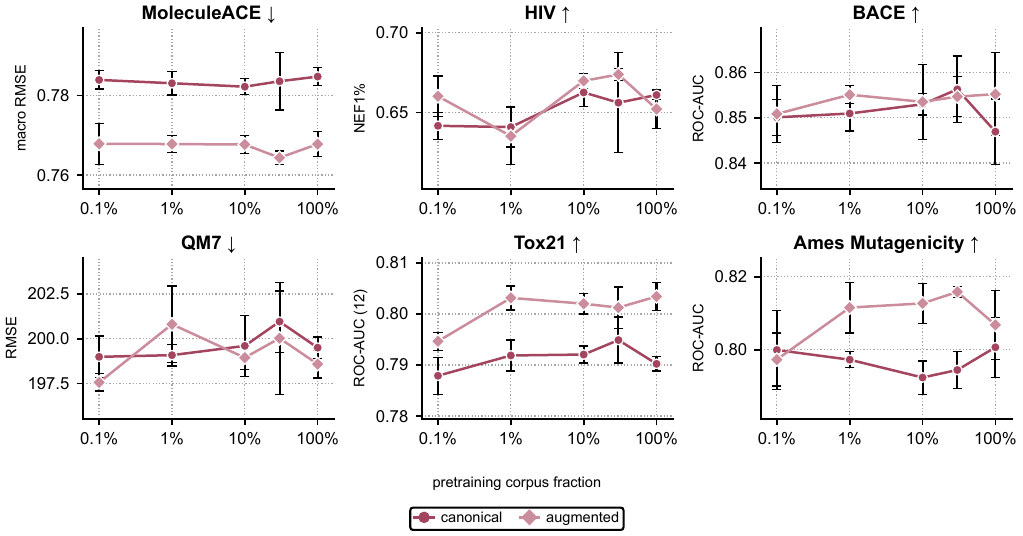}
    \caption{\textbf{Canonical versus enumerated SMILES in pretraining.}
    Canonical and enumerated pretraining are matched at five corpus
    fractions $\times$ 3 pretraining seeds.}
    \label{SI_fig_d_canon_vs_enum}
\end{figure}

We find no strong differences in performance although the augmented pretraining procedure edges out a win in MoleculeACE and Tox21, see Figure \ref{SI_fig_d_canon_vs_enum}. In absolute terms, the difference in performance is relatively minor and the augmented model is only slightly more robust towards non-canonical input SMILES, see also Figure \ref{fig_G}. In summary, pretraining on augmented SMILES is certainly the safer choice and may have small benefits on certain datasets, although downstream performance is much more sensitive towards the choice of pretraining objective and dataset.

\section{Controlling for weight-conditioning in unsupervised pretraining}
\label{sec:SI_weights_conditioning}

The results in Figure \ref{fig_E}b in the main text establish that the benefits of unsupervised pretraining can largely be recovered even by training on corrupted SMILES. To distinguish the effects of learning generalizable features from optimization and improved initialization, we ran a control experiment \cite[cf.][]{raghu2019transfusion, lu2022frozen, wu2022insights}. We opted to construct control models that appear numerically similar to a trained model while not having undergone any training. We achieve this either simply by resampling each weight matrix from its own per-tensor mean and variance; or by randomly permuting  matrix entries. The LayerNorm values are copied verbatim from the trained model. Table~\ref{tab:pretraining_decomposition} reports the decomposition.

\begin{table}[h!]
\centering
\footnotesize
\setlength{\tabcolsep}{4.5pt}
\caption{\textbf{Controlling for weight conditioning in unsupervised pretraining.} 
We report the gain in performance over a randomly initialized encoder (lift). \emph{Weight conditioning} refers to the weight-matched control models (mean of 3 seeds). \emph{Generic} is gain attributable to the Wikipedia arm; \emph{chemistry} is the gain due to using real SMILES; both with the effect of conditioning subtracted.}
\label{tab:pretraining_decomposition}
\begin{tabular}{@{}l cc cc cc @{\hspace{6pt}}||@{\hspace{6pt}} c@{}}
\toprule
& \multicolumn{2}{c}{\emph{measured lift}}
& \multicolumn{2}{c}{\emph{weight conditioning}}
& \multicolumn{2}{c}{\emph{attributable to}} & \\
\cmidrule(lr){2-3} \cmidrule(lr){4-5} \cmidrule(lr){6-7}
\textbf{Panel} & \textbf{real} & \textbf{wiki} & \textbf{real} & \textbf{wiki}
& \textbf{generic} & \textbf{chemistry} & \textbf{Chemistry\ share} \\
\midrule
QM7          & $+2.86$  & $+2.84$  & \textbf{2.74} & \textbf{3.12} & $-0.28$        & $+0.40$        & --- \\
MoleculeACE  & $+6.39$  & $+5.94$  & 0.83          & 1.44          & \textbf{$+4.50$} & $+1.06$      & 19\% \\
HIV          & $+22.44$ & $+15.60$ & 5.42          & 2.75          & \textbf{$+12.85$} & $+4.17$     & 25\% \\
BACE         & $+6.45$  & $+1.94$  & 0.78          & 0.97          & $+0.97$        & \textbf{$+4.70$} & 83\% \\
Tox21        & $+3.24$  & $+0.44$  & 0.26          & $-0.13$       & $+0.57$        & \textbf{$+2.41$} & 81\% \\
Ames         & $+3.78$  & $-0.10$  & 1.07          & 0.93          & $-1.03$        & \textbf{$+3.74$} & 138\% \\
\bottomrule
\end{tabular}

\vspace{2pt}
\end{table}

Three mechanisms are separable, and their relative sizes are task-dependent: On QM7 for example, the control reproduces the full effect (2.74 of 2.86 for the real corpus, 3.12 of 2.84 for Wikipedia). The apparent pretraining benefits on this task seemingly come down to better weight initialization and no learned chemistry. On MoleculeACE and HIV conditioning accounts for only 13--24\%, and a corpus containing no chemistry supplies 75--81\% of what real SMILES supplies, suggesting that for these tasks low-level corpus statistics are most useful. On BACE, Tox21 and Ames the ordering reverses: 81--138\% of the improvements only derive from real SMILES, and on Ames a non-chemical corpus is actively harmful. These tasks seemingly require true, chemically correct and representative token sets. Future work is needed  to better understand what type of pretraining information is most useful for which type of task in unsupervised pretraining. 



\end{document}